\documentclass[aps,prr,twocolumn,superscriptaddress]{revtex4-1}
\usepackage{amsfonts}
\usepackage{graphicx}
\usepackage{epstopdf}
\usepackage{epsfig}
\usepackage{amsmath}
\usepackage[normalem]{ulem}
\usepackage{multirow}
\usepackage{color}
\usepackage{makecell}
\usepackage{array}
\usepackage{booktabs}
\usepackage{mathtools}
\usepackage{bm}
\usepackage{color}
\usepackage{array}
\usepackage{booktabs}
\usepackage{tabularx}

\newcolumntype{Y}{>{\centering\arraybackslash}X}

\newcommand{\Rmnum}[1]{\expandafter\@slowromancap\romannumeral #1@}
\makeatother

\begin{document}

\title{Adaptable Hamiltonian neural networks}

\author{Chen-Di Han}
\affiliation{School of Electrical, Computer and Energy Engineering, Arizona State University, Tempe, Arizona 85287, USA}

\author{Bryan Glaz}
\affiliation{Vehicle Technology Directorate, CCDC Army Research Laboratory, 2800 Powder Mill Road, Adelphi, MD 20783-1138, USA}

\author{Mulugeta Haile}
\affiliation{Vehicle Technology Directorate, CCDC Army Research Laboratory, 2800 Powder Mill Road, Adelphi, MD 20783-1138, USA}

\author{Ying-Cheng Lai} \email{Ying-Cheng.Lai@asu.edu}
\affiliation{School of Electrical, Computer and Energy Engineering, Arizona State University, Tempe, Arizona 85287, USA}
\affiliation{Department of Physics, Arizona State University, Tempe, Arizona 85287, USA}

\begin{abstract}

The rapid growth of research in exploiting machine learning to predict chaotic systems has revived a recent interest in Hamiltonian Neural Networks (HNNs) with physical constraints defined by the Hamilton's equations of motion, which represent a major class of physics-enhanced neural networks. We introduce a class of HNNs capable of adaptable prediction of nonlinear physical systems: by training the neural network based on time series from a small number of bifurcation-parameter values of the target Hamiltonian system, the HNN can predict the dynamical states at other parameter values, where the network has not been exposed to any information about the system at these parameter values. The architecture of the HNN differs from the previous ones in that we incorporate an input parameter channel, rendering the HNN parameter--cognizant. We demonstrate, using paradigmatic Hamiltonian systems, that training the HNN using time series from as few as four parameter values bestows the neural machine with the ability to predict the state of the target system in an entire parameter interval. Utilizing the ensemble maximum Lyapunov exponent and the alignment index as indicators, we show that our parameter-cognizant HNN can successfully predict the route of transition to chaos. Physics-enhanced machine learning is a forefront area of research, and our adaptable HNNs provide an approach to understanding machine learning with broad applications.

\end{abstract}

\date{\today}

\maketitle

\section{Introduction} \label{sec:intro}

A daunting challenge in machine learning is the lack of understanding of the
inner working of the artificial neural networks. As machine learning has been
increasingly incorporated into many vital structures and systems that support
the functioning of the modern society, it is imperative to develop a general
understanding of the inner gears of the underlying neural networks. For
example, feed-forward neural networks or multilayer perceptrons constitute
the fundamentals of modern deep learning machines with broad applications in
image, video and audio processing~\cite{lecun2015deep}. Such a neural machine
typically consists of an input layer, a large number of hidden layers, and
an output layer. From the input layer on, nodes in the same layer do not
interact with each other, but they are connected with the nodes in the next
layer via a set of weights and biases whose values are determined through
training, where the paradigmatic method of stochastic gradient descent
(SGD)~\cite{Goodfellow-et-al-2016} is often used. How the networks in
different layers work together to solve a specific problem remains unknown.
In another line of research, reservoir computing, a class of recurrent neural
networks~\cite{Jaeger:2001,MNM:2002,JH:2004,MJ:2013}, has
gained considerable momentum since 2017 as a powerful paradigm for
model-free, fully data driven prediction of nonlinear and chaotic dynamical
systems~\cite{HSRFG:2015,LBMUCJ:2017,PLHGO:2017,LPHGBO:2017,DBN:book,LHO:2018,PWFCHGO:2018,PHGLO:2018,Carroll:2018,NS:2018,ZP:2018,WYGZS:2019,GPG:2019,JL:2019,VPHSGOK:2019,FJZWL:2020,ZJQL:2020}.
A reservoir computing machine constitutes an input layer, a single hidden layer,
and an output layer. Differing from the network structure of a multilayer
perceptron, the network in the hidden layer of a reservoir computing machine
has a complex topology in which the nodes are coupled with each other following
some probability distribution. Another difference is that, in feed-forward
neural networks, only the weights and biases connecting the hidden layer and
the output layer neurons are determined by training, while in reservoir
computing those parameters as well as the weights of the complex network in
the hidden layer are pre-defined. A well trained reservoir machine can
generate accurate prediction of the state evolution of a chaotic system for
a duration that is typically several times longer than that which can be
achieved using the traditional methodologies in nonlinear time series analysis.
This is remarkable, considering the hallmark of chaos: sensitive dependence on
initial conditions, which rules out long-term prediction. Yet, there is little
understanding of how the internal network dynamics of reservoir computing
machines behave or ``manage'' to replicate accurately (for some amount of
time) the chaotic evolution of the true system.

At the present, to develop a general explainable framework to encompass
various types of machine learning is not feasible. In this regard, a promising
direction of pursuit is the so-called physics-enhanced machine learning,
in which the neural networks are designed to solve specific physics problems
with the goal to enhance the learning efficiency through exploiting the
underlying physical principles or constraints. The idea was articulated almost
three decades ago~\cite{de1993class}, when the principle of Hamiltonian
mechanics was incorporated into the design of neural networks, leading to {\em
Hamiltonian Neural Networks} (HNNs) that have recently gained renewed
attention~\cite{greydanus2019hamiltonian,toth2019hamiltonian,
bertalan2019learning,choudhary2019physics,garg2019hamiltonian}. Comparing with
traditional neural networks, in an HNN, the energy is conserved. It has been
demonstrated that, an HNN can be trained to possess the power to predict the
dynamical evolution of the target Hamiltonian system in both integrable and
chaotic regimes, provided that the network is trained with data taken from the
same set of parameter values at which the prediction is to be
made~\cite{greydanus2019hamiltonian,toth2019hamiltonian,
bertalan2019learning,choudhary2019physics,garg2019hamiltonian}. Recently the
principle of HNN has been generalized~\cite{chen2019symplectic} to systems
described by the Lagrangian equation of motion~\cite{cranmer2020lagrangian} and
a general type of ordinary differential equations~\cite{sanchez2019hamiltonian}
or coordinate transforms~\cite{dulberg2020learning,choudhary2020forecasting}
with applications in robotics~\cite{lutter2019deep,havoutis2010geodesic}.

In this paper, we address adaptability, a fundamental issue in machine
learning, of Hamiltonian neural networks. More precisely, we consider the
situation where a target Hamiltonian system can experience slow drift or
sudden changes in some parameters. Slow environmental variations can lead to
adiabatic parameter drifting, while external disturbances can lead to sudden
parameter changes. We ask if it is possible to design HNNs, which are trained
with data from a small number of parameter values of the target system, to
have the predictive power for parameter values that are not in the training
set. Inspired by the recent work on predicting critical transitions and
collapse in dissipative dynamical systems based on reservoir
computing~\cite{falahian2015artificial,cestnik2019inferring,kim2020teaching,klos2020dynamical,KFGL:2021},
we articulate a class of HNNs whose input layer contains a set of channels that
are specifically used for inputting the values of the distinct parameters of
interest to the neural network. The number of the parameter channels is equal
to the number of freely varying parameters in the target Hamiltonian system.
The simplest case is where the target system has a single bifurcation or
control parameter so only one input parameter channel to the neural network is
necessary. We demonstrate that, by incorporating such a parameter channel into
a feed-forward type of HNNs and conducting training using time series data
from a small number of bifurcation parameter values (e.g., four), we
effectively make the HNN adaptable to parameter variations. That is, the
so-trained HNN has inherited the rules governing the dynamical evolution of
the target Hamiltonian system. When a parameter value of interest, which is
not in the training parameter set, is fed into the HNN through the parameter
channel, the machine is capable of generating dynamical behaviors that
statistically match those of the target system at this particular parameter
value. The HNN has thus become adaptable because it has never been exposed
to any information or data from the target system at this parameter value,
yet the neural machine can reproduce the dynamical behavior. Using the
H\'{e}non-Heiles model as a prototypical target Hamiltonian system, we
demonstrate that our adaptable HNN can successfully predict the dynamical
behaviors, integrable or chaotic, for any parameter values that are reasonably
close to those in the training parameter set. Remarkably, by feeding a
systematically varying set of bifurcation parameter values into the parameter
channel, the HNN can successfully predict the transition to chaos in the target
Hamiltonian system, which we characterize using two measures: the ensemble
maximum Lyapunov exponent and the alignment index. It is worth emphasizing
that, in the existing literature on
HNNs~\cite{greydanus2019hamiltonian,toth2019hamiltonian,bertalan2019learning,choudhary2019physics,garg2019hamiltonian},
training and prediction are done at the same set of parameter values of the
target Hamiltonian system, but our work goes beyond by making the HNN
significantly more powerful with enhanced and expanded predictability.

We remark that, in physics, machine learning has been exploited to solve
difficult problems in particle physics~\cite{guest2018deep,radovic2018machine},
quantum many-body systems~\cite{carleo2017solving}, inverse design in optical
systems~\cite{peurifoy2018nanophotonic}, and quantum
information~\cite{dunjko2018machine,carleo2019machine}. However, the working
mechanisms of the underlying neural networks remain largely
unknown~\cite{iten2020discovering}. The physics enhanced HNNs studied
here are different from these applications, as we focus on exploiting
physical principles to enable neural networks with unprecedented predictive
power with respect to parameter variations.

In Sec.~\ref{sec:system}, we describe the architecture of the articulated
parameter-cognizant HNNs and the method of training. In
Sec.~\ref{sec:result_1}, we present results of predicting the dynamical
behavior of the H\'{e}non-Heiles system in a wide parameter region, including
the transition to chaos based on calculating the ensemble maximum Lyapunov
exponent and the minimum alignment index. In general, the prediction accuracy
depends on how ``close'' the desired parameter value is to the training
regime. In Sec.~\ref{sec:issues}, we address a number of pertinent issues
such as the choosing of the training parameter values, multiple
parameter channels, and HNNs for a Hamiltonian system defined by the
one-dimensional Morse potential. A summarizing discussion and speculations
are offered in Sec.~\ref{sec:discussion}.

\section{Parameter-cognizant Hamiltonian neural networks}\label{sec:system}

The central idea for physics-enhanced machine learning is to ``force'' the
dynamical evolution of the neural network to follow certain physical rules
or constraints, examples of which are Hamilton's equations of
motion~\cite{greydanus2019hamiltonian,toth2019hamiltonian,bertalan2019learning,
choudhary2019physics,garg2019hamiltonian}, Lagrangian
equations~\cite{cranmer2020lagrangian}, or the principle of least
action~\cite{karkar2020principle,yang2020learning}.
In particular, the structure of HNNs is such that the underlying neural
dynamical system is effectively a Hamiltonian system for which the energy
is conserved during the evolution. Different from previous
work~\cite{greydanus2019hamiltonian,toth2019hamiltonian,bertalan2019learning,
choudhary2019physics,garg2019hamiltonian}, the bifurcation parameter of the
target Hamiltonian system serves as an input ``variable'' to the neural
network through an additional input channel so that the HNN learns to
associate the input time series with the specific value of the bifurcation
parameter. Using time series from a small number of distinct bifurcation
parameter values to train the HNN, it can gain the ability to ``sense'' the
changes in the dynamics (or dynamical ``climate'') of the target system with
the bifurcation parameter.

\begin{figure} [ht!]
\centering
\includegraphics[width=\linewidth]{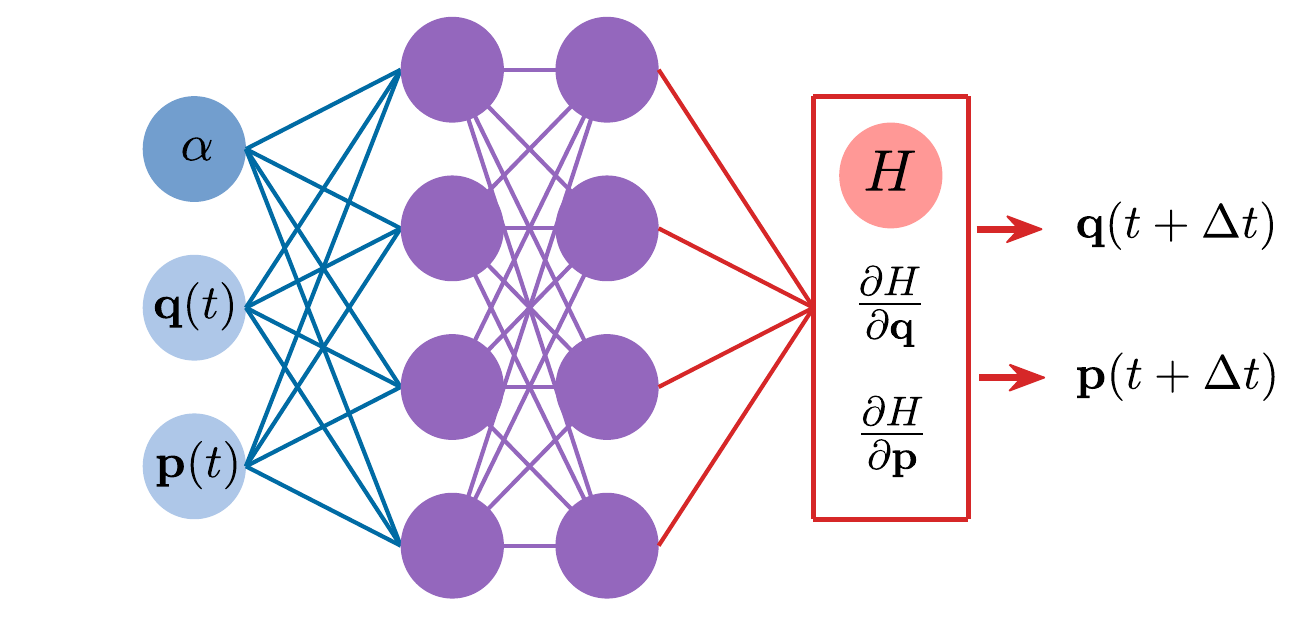}
\caption{Structure of parameter-cognizant HNN. The input channels are denoted
by the blue circles, which are connected to the first hidden layer (purple
circles). The blue circle denoted by ``$\alpha$'' is the parameter input
channel that feeds the value of the bifurcation parameter of the target
Hamiltonian system, together with the time series $\mathbf{q}(t)$ and
$\mathbf{p}(t)$ through the corresponding input channels, into the first
hidden layer. There are two hidden layers. The output variables are the
partial derivatives of the Hamiltonian of the target system with respect to
the canonical coordinates and momenta, together with the Hamiltonian, which
determine the dynamical state at the next time step.}
\label{fig:HNN}
\end{figure}

The structure of our articulated parameter-cognizant HNN is shown in
Fig.~\ref{fig:HNN}, where the input contains three parts: the position and
momentum variables of the target system, and the bifurcation parameter. To be
concrete, we use two hidden layers, where each layer contains $200$
artificial neurons (nodes). The third layer is the output, which contains
a single node whose dynamical state corresponds to the Hamiltonian of the
target system. Let $\mathbf{y}$ denote the set of dynamical variables of each
layer. The transform from the dynamical variables in the $i$th layer to those
in the $(i+1)$th layer follows the following rule:
\begin{equation} \label{eq:1_layer}
	\mathbf{y}^{i+1}=\bm{\sigma}^i(\mathcal{W}^{i}\cdot \mathbf{y}^{i} + \mathbf{b}^{i}),
\end{equation}
where $\bm{\sigma}^i$ is a given nonlinear activation function, $\mathcal{W}^i$
is the weight matrix and $\mathbf{b}^i$ is bias vector associated with the
neurons in the $i$th layer, which are to be determined through training. We
set the output as the spatial derivatives of the input variables to force the
dynamics of the neural network to follow the Hamilton's equations of motion.
The derivatives are calculated through the back prorogation algorithm. Once
the output is known, the loss function defined as
\begin{equation} \label{eq:loss}
\mathcal{L}=\left\| \frac{\partial H}{\partial \mathbf{q}}+\frac{d\mathbf{p}_\text{real}}{dt}\right\|+\left\| \frac{\partial H}{\partial \mathbf{p}}-\frac{d\mathbf{q}_\text{real}}{dt}\right\|,
\end{equation}
can be calculated. Through the training process, we optimize the weights and
biases in Eq.~\eqref{eq:1_layer} by minimizing the loss function. This is
done by using the standard SGD method~\cite{Goodfellow-et-al-2016}. The whole
process from network construction and training to carrying out the prediction
is accomplished by using the open source package Tensorflow and
Keras~\cite{abadi2016tensorflow,chollet2015keras}.

Table~\ref{ta:0_algorithm} summarizes the structure and parameters of our
HNN. It has about 40,000 unknown parameters to be optimally determined through
training. The computation can be quite efficient even without using parallel
or GPU acceleration. The anticipation is that, after training with time series
data from a small number of distinct bifurcation parameter values, the HNN can
predict the dynamical behavior of the target Hamiltonian system in a wide
parameter interval, where the parameter variations are implemented through the
input parameter channel to the HNN.

\begin{table}
\caption{Specifications of HNN}
\label{ta:0_algorithm}
\begin{tabularx}{\linewidth}{YY}
\hline\hline
\specialrule{0em}{1pt}{1pt}
Description & Values \\
\specialrule{0em}{1pt}{1pt}
\hline
\specialrule{0em}{1pt}{1pt}
Number of hiden layers &  $2$ \\
\specialrule{0em}{1pt}{1pt}
Neurons per layer &  $200$ \\
\specialrule{0em}{1pt}{1pt}
Optimizer &  Adam \\
\specialrule{0em}{1pt}{1pt}
Epochs &  $500$ \\
\specialrule{0em}{1pt}{1pt}
Activation functioin &  tanh \\
\specialrule{0em}{1pt}{1pt}
\hline\hline
\end{tabularx}
\end{table}

\begin{figure} [ht!]
\centering
\includegraphics[width=\linewidth]{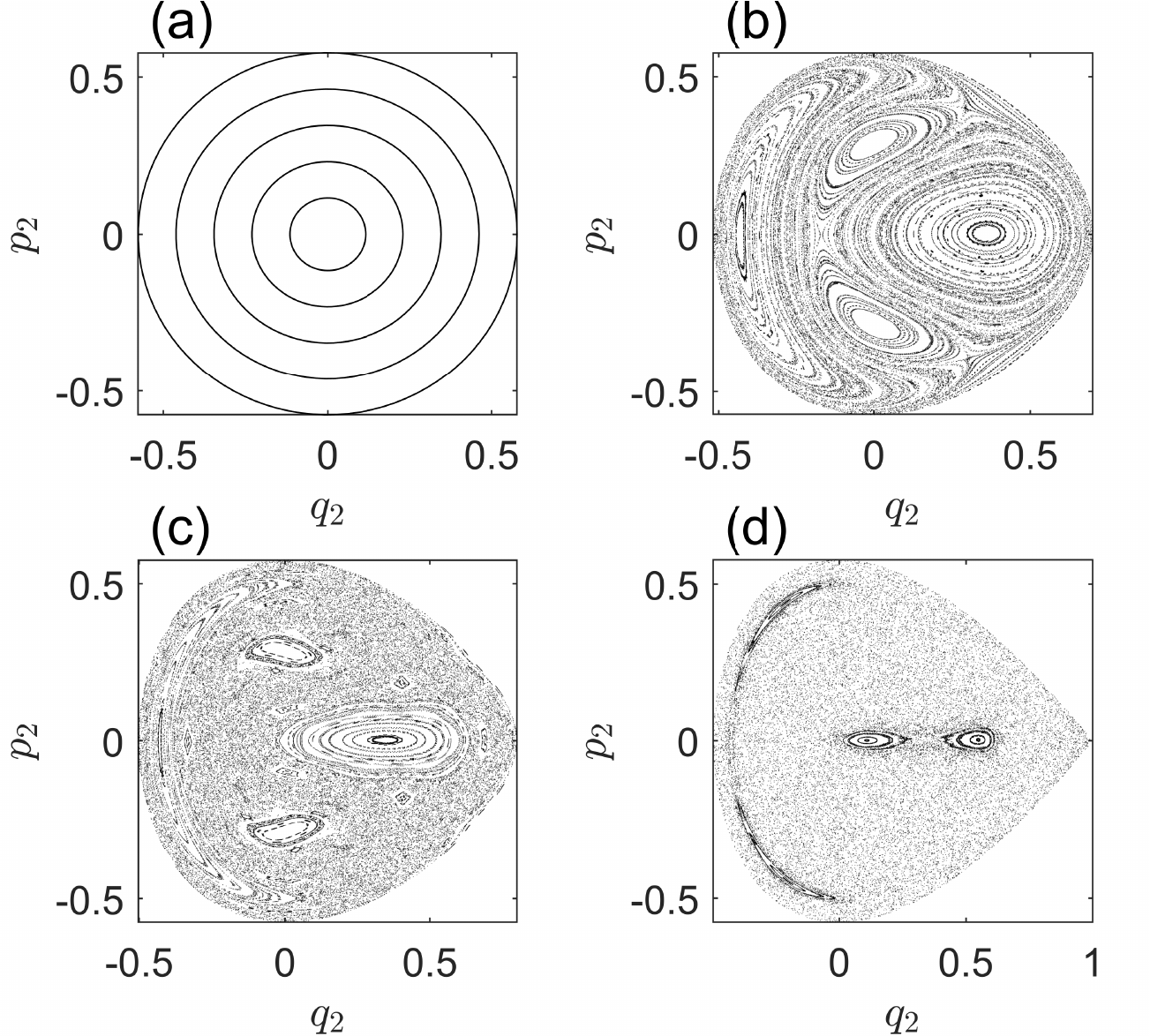}
\caption{Distinct types of dynamical behaviors in the H\'{e}non-Heiles system.
The energy value is $E = 1/6$. (a-d) Integrable, mixed, small and largely
chaotic dynamics for $\alpha = 0$, 0.7, 0.9, and 1.0, respectively, on the
Poincar\'{e} surface of section defined by $q_1 = 0$.}
\label{fig:HH_1}
\end{figure}

\section{Adaptable Hamiltonian neural networks for predicting transition to chaos} \label{sec:result_1}

To test the adaptability of our parameter-cognizant HNN for predicting state
evolution and dynamical transitions in Hamiltonian systems, we use the
paradigmatic H\'{e}non-Heiles model~\cite{henon1964applicability}. It is a
two-degrees-of-freedom system for investigating distinct types of Hamiltonian
dynamics including integrable, mixed, and chaotic behaviors and the transitions
among them. It was originated from the gravitational three-body
system~\cite{henon1964applicability}, with applications in contexts such as
molecular dynamics~\cite{waite1981mode,feit1984wave,vendrell2011multilayer}.

\subsection{System description} \label{subsec:HH_system}

The H\'{e}non-Heiles Hamiltonian is
\begin{equation} \label{eq:Hamiltonian}
H=\frac{1}{2}\left(p_1^2+p_2^2\right) + \frac{1}{2}\left(q_1^2+q_2^2 \right)+\alpha\left(q_1^2q_2-\frac{1}{3}q_2^3\right),
\end{equation}
where $q_1$ and $q_2$ denote the coordinates, $p_1$ and $p_2$ are the
corresponding momenta, $\alpha > 0$ is a bifurcation parameter that sets the
magnitude of the nonlinear potential function describing, e.g., the
dissociation energy in molecules~\cite{waite1981mode,feit1984wave,
vendrell2011multilayer}. The dynamics of system Eq.~\eqref{eq:Hamiltonian}
depend not only on $\alpha$, but also on the energy $E$ of the system that
is conserved during the dynamical evolution. The maximum value of the
potential function is $E_\text{max}=1/(6\alpha^2)$. For $\alpha = 0$,
$E_\text{max}$ diverges, so all trajectories are bounded. For $\alpha > 0$,
if the particle energy exceeds $E_\text{max}$, the Hamiltonian system becomes
open with scattering trajectories that can escape to infinity. To train the
adaptable HNN, bounded trajectories are required, so we set $0 \le \alpha \le 1$
and $E \le 1/6$. (For particle energy above the threshold, chaotic scattering
dynamics and fractal geometry can arise~\cite{de1999fractal,
seoane2006basin,seoane2007fractal}.) As the value of $\alpha$ increases from
zero to one, characteristically different dynamical behaviors can arise, such
as integrable, mixed, and chaotic. In particular, for $\alpha = 0$, the
nonlinear term in Eq.~\eqref{eq:Hamiltonian} disappears and the system
becomes a harmonic oscillator - an integrable system. In this case, the
entire phase space contains periodic and quasiperiodic orbits only, as shown
in Fig.~\ref{fig:HH_1}(a). As $\alpha$ increases from zero, the system becomes
nonlinear and chaotic seas amid the Kol'mogorov-Arnol'd-Moser (KAM) islands
can arise in the phase space, giving rise to mixed dynamics, as shown in
Fig.~\ref{fig:HH_1}(b) for $\alpha = 0.7$ and $E = 1/6$. For $\alpha=0.9$,
$\alpha = 1$ and $E = 1/6$, most trajectories in the phase space are chaotic,
as shown in Figs.~\ref{fig:HH_1}(c) and \ref{fig:HH_1}(d).

\subsection{Training and testing of adaptability} \label{subsec:HH_TTA}

Our goal of training is to ``instill'' certain adaptable power into the HNN.
To achieve this, we choose a number of distinct values of the bifurcation
parameter $\alpha$. For each $\alpha$ value, we randomly choose initial
conditions with energy below the escape threshold $E_\text{max}$ and
numerically integrate the Hamilton's equations of motion to generate particle
trajectories in the phase space. Because of the mixed dynamics, the training
data contain both integrable and chaotic orbits. Specifically, the time
interval of the trajectory is $0 \le t\le 1000$, which contains hundreds of
oscillation cycles, and we collect training data using the sampling time step
$dt=0.1$. The energy associated with the training data is maintained to be
constant to within $10^{-6}$.

In general, the weights and biases of the adaptable HNN determined by the SGD
method depend on the training data set. To reduce the prediction error,
an ensemble of HNNs can be used~\cite{KFGL:2021}. Concretely, for
each value of $\alpha$, we generate $20$ different sets of data for training,
leading to an ensemble of 20 HNNs. The parameter setting for training is
listed in Tab.~\ref{ta:1_data}.

\begin{table}
\caption{List of training parameters for H\'{e}non-Heiles system}
\label{ta:1_data}
\begin{tabularx}{\linewidth}{YY}
\hline\hline
\specialrule{0em}{1pt}{1pt}
Description  &  Values\\
\specialrule{0em}{1pt}{1pt}
\hline
\specialrule{0em}{1pt}{1pt}
Neural network ensembles &  $20$ \\
\specialrule{0em}{1pt}{1pt}
Energy samples &  $7$ \\
\specialrule{0em}{1pt}{1pt}
Orbit per energy & $1$\\
\specialrule{0em}{1pt}{1pt}
Orbit length &  $1000$ \\
\specialrule{0em}{1pt}{1pt}
Time step &  $0.1$ \\
\specialrule{0em}{1pt}{1pt}
Training parameter set &  $\alpha\in\{0.2, 0.4, 0.6, 0.8\}$ \\
\specialrule{0em}{1pt}{1pt}
\hline\hline
\end{tabularx}
\end{table}

After the training, all the weights and biases in Eq.~\eqref{eq:1_layer} are
determined. The Hamiltonian and its derivatives for each network in the
HNN ensemble can be evaluated for any input, leading to the average derivative
values. To characterize the prediction accuracy for different values of the
bifurcation parameter, we use the root-mean square error (RMSE) that can
be calculated from the difference between the HNN predicted and the true
orbits. For $\alpha=0$, the motion is integrable so the predicted orbit is
always close to some real orbit, leading to exceedingly small errors. In this
case, we take advantage of one feature of HNNs that it directly yields the
Hamiltonian function, from which the potential function can be calculated.
It is thus convenient to use the relative error between the predicted
potential function and the true one to characterize the HNN performance,
which is defined as
\begin{equation} \label{eq:DV}
\langle \Delta V\rangle \equiv \frac{\overline{|V_\text{pred}-V_\text{real}|}}{\overline{V_\text{real}}},
\end{equation}
where the average is taken in the region of $V_\text{real}<1/6$. The predicted
potential profile is given by $V_\text{pred}=H_\text{pred}-C$, where
$C=\text{min}(H_\text{pred})$ so that the minimum value of $V_\text{pred}$
is zero. Note that the average in Eq.~(\ref{eq:DV}) is calculated from
an integral in a 2D domain in the physical space, for which the boundary of
the domain needs to be specified. A natural choice of the criterion to set
the boundary would be $V(x,y) < E_\text{max} = 1/6$, but occasionally the
predicted orbit will diverge. Numerically, there are different ways to overcome
this difficulty. For example, if the boundary is set according to the
criterion: $\mbox{max}(V_\text{pred},V_\text{real})<1/6$, then almost all
orbits are bounded, rendering calculable the error $\langle \Delta V\rangle$.

We demonstrate that HNN can be used to reconstruct the Hamiltonian of the 
target system. Consider phase-space points for $H(\alpha,q_1,q_2,p_1,p_2)<1/6$ 
and expand the Hamiltonian about the origin using the Taylor series:
\begin{equation}
H_\text{pred}(\alpha)=\sum_{i_1,i_2,i_3,i_4}\beta_{i_1i_2i_3i_4}(\alpha)q_1^{i_1}q_2^{i_2}p_1^{i_3}p_2^{i_4},
\end{equation}
where $\beta$'s are the expansion coefficients, and the sum is taken according
to $0\leq \text{sum} (i_1,i_2,i_3,i_4)\leq 3$, which contains in total $35$ 
terms. Comparing with true Hamiltonian Eq.~\eqref{eq:Hamiltonian}, only six 
terms are non-zero. 

\begin{figure} [ht!]
\centering
\includegraphics[width=\linewidth]{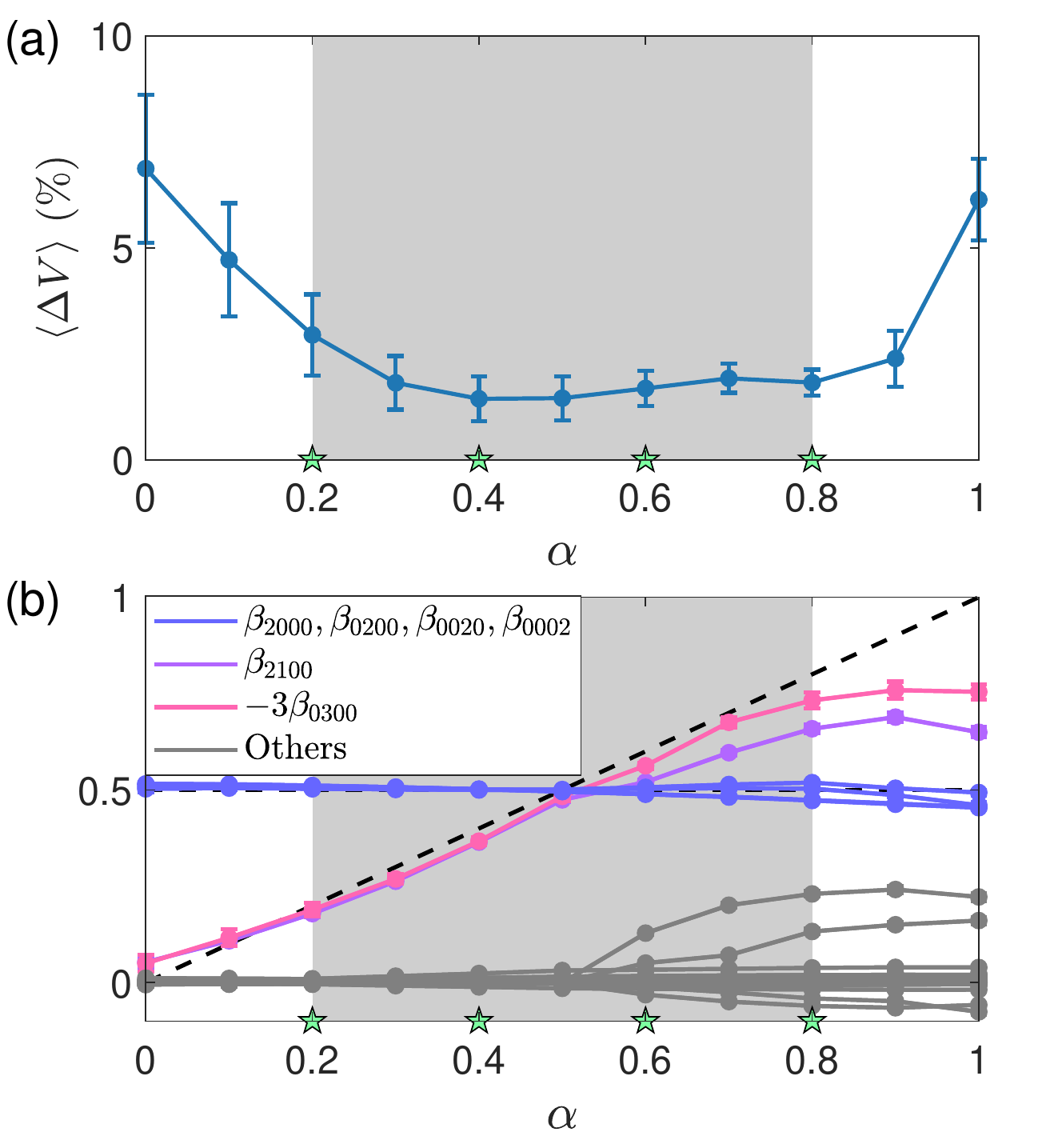}
\caption{ 
(a) Relative error in predicting the potential function of the
H\'{e}non-Heiles system. Shown is the error versus the bifurcation parameter
$\alpha$ in the unit interval. The HNN is trained with time series data from
four values of $\alpha$: $\alpha\in\{0.2, 0.4, 0.6, 0.8\}$ (the four green
pentagons). In the shaded region that contains these four values of $\alpha$,
the relative error is less than $2\%$, demonstrating the
adaptability of the HNN in predicting the target Hamiltonian system for
parameter values not in the training set. The adaptability extends even
outside the shaded region but with larger error (still within $8\%$ though). 
(b) Coefficients of the Taylor expansion for $H_\text{pred}$ versus the
bifurcation parameter $\alpha$, where $\beta_{2000}$, $\beta_{0200}$, 
$\beta_{0020}$ and $\beta_{0002}$ correspond to the first four square terms 
in Eq.~\eqref{eq:Hamiltonian} whose true value is $1/2$, and $\beta_{2100}$ and 
$\beta_{0300}$ correspond to the two cubic terms that are proportional to 
$\alpha$. Other terms in the expansion do not appear in the original 
Hamiltonian, among which the first two largest ones are $\beta_{3000}$ 
and $\beta_{1200}$ that correspond to other cubic potential terms.}
\label{fig:HH_2}
\end{figure}

We train the HNN at four values of the bifurcation parameter:
$\alpha\in\{0.2, 0.4, 0.6, 0.8\}$. For each $\alpha$ value, we choose seven
random initial conditions with their energies below the threshold.
Figure~\ref{fig:HH_2}(a) shows the relative error in predicting the potential
function for $\alpha\in (0,1)$. The interval in $\alpha$ can be divided into
two parts: the shaded region $\alpha\in [0.2,0.8]$ that contains the four
values of $\alpha$ used in training, and the blank regions on both side of
the shaded region. In the shaded region, the relative error is less than
$2\%$, but the error increases away from the shaded region. 
Figure~\ref{fig:HH_2}(b) shows the expansion coefficients for the predicted 
Hamiltonian. Comparing with the terms in the real Hamiltonian, our HNN 
predicts accurately the linear terms. For the nonlinear terms, the HNN 
reproduces the behavior with the variation in the bifurcation parameter 
$\alpha$, where the errors are small in the shaded region in 
Fig.~\ref{fig:HH_2}(b) but relatively large outside the region.

\begin{figure} [ht!]
\centering
\includegraphics[width=\linewidth]{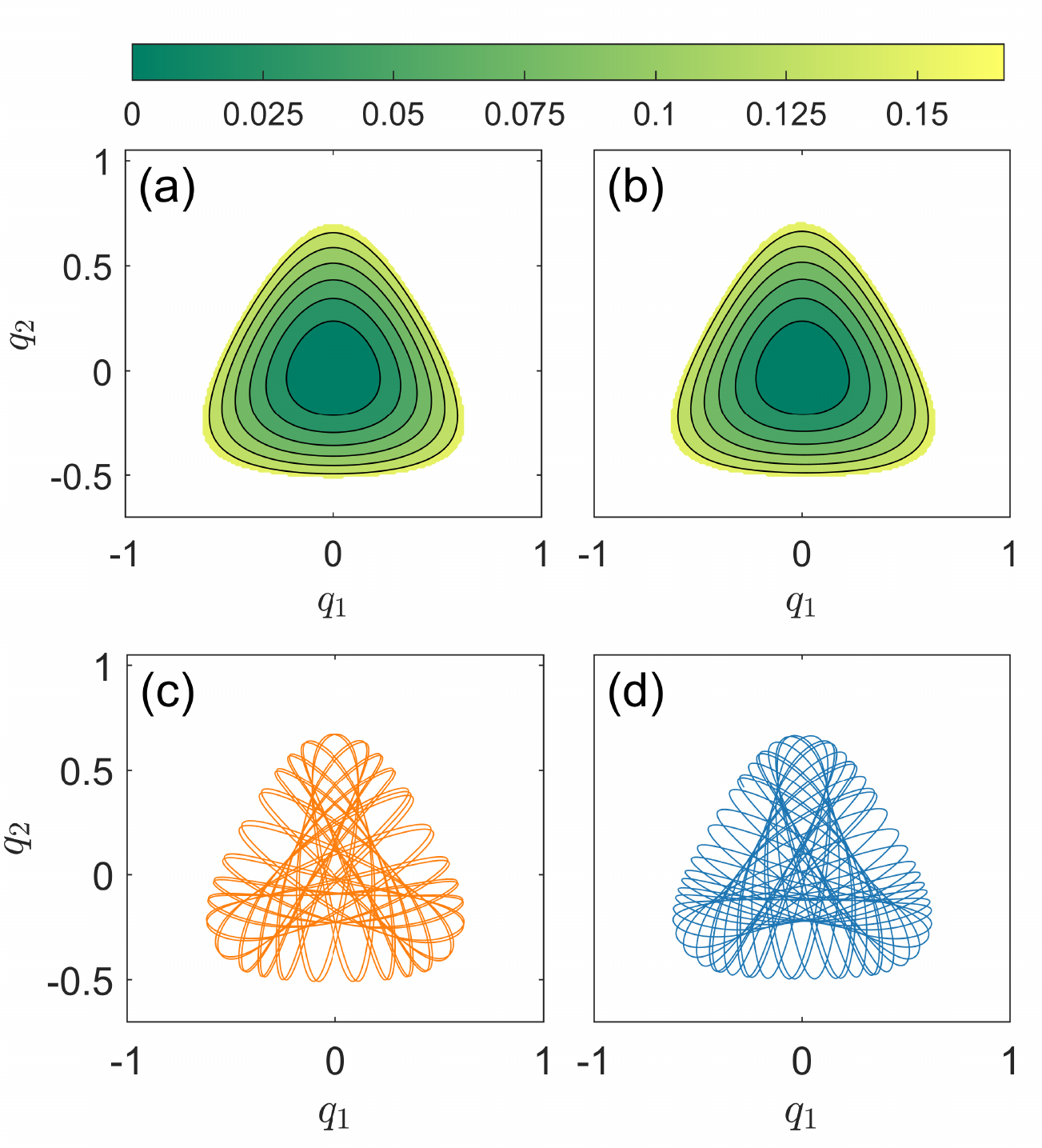}
\caption{Testing the adaptability of HNN for parameter values in between
two training points. (a,b) True and predicted contour maps of the potential
function for $\alpha=0.7$, respectively, where the latter is obtained by
extracting the Hamiltonian at different positions with constant momentum input
with normalization. (c,d) True and predicted orbits from the initial
condition $[q_1,q_2,p_1,p_2]=[0,0,1/\sqrt{6},1/\sqrt{6}]$, respectively,
which are quasiperiodic.}
\label{fig:HH_3}
\end{figure}

To examine the adaptability of our parameter-cognizant HNN in more detail, we
take, for example, $\alpha=0.7$ in between the two training points $\alpha=0.6$
and $\alpha=0.8$, for which the vast majority of the orbits with energy $E=1/6$
are quasiperiodic, as can be seen from Fig.~\ref{fig:HH_1}(b).
Figures~\ref{fig:HH_3}(a) and \ref{fig:HH_3}(b) show the true and predicted
potential functions for $E<1/6$, respectively, which are essentially
indistinguishable. Figures~\ref{fig:HH_3}(c) and \ref{fig:HH_3}(d) show
some representative true and predicted orbits starting from the same initial
condition, which agree with each other qualitatively but differ in detail.
Particularly worth emphasizing is the fact that for the predicted quasiperiodic
orbit, the energy can be maintained at a constant value. In fact, we have
tested the method of reservoir computing~\cite{HSRFG:2015,LBMUCJ:2017,PLHGO:2017,LPHGBO:2017,DBN:book,LHO:2018,PWFCHGO:2018,PHGLO:2018,Carroll:2018,NS:2018,ZP:2018,WYGZS:2019,GPG:2019,JL:2019,VPHSGOK:2019,FJZWL:2020,ZJQL:2020}
for predicting the orbit and find that, while it typically yields a more
accurate orbit in short time (e.g., a few cycles), in the long run the
energy is not conserved and the prediction error becomes large. Overall, since
the testing bifurcation parameter value $\alpha = 0.7$ is sandwiched between
two training points, our parameter-cognizant HNN exhibits a strong adaptability.

\begin{figure} [ht!]
\centering
\includegraphics[width=\linewidth]{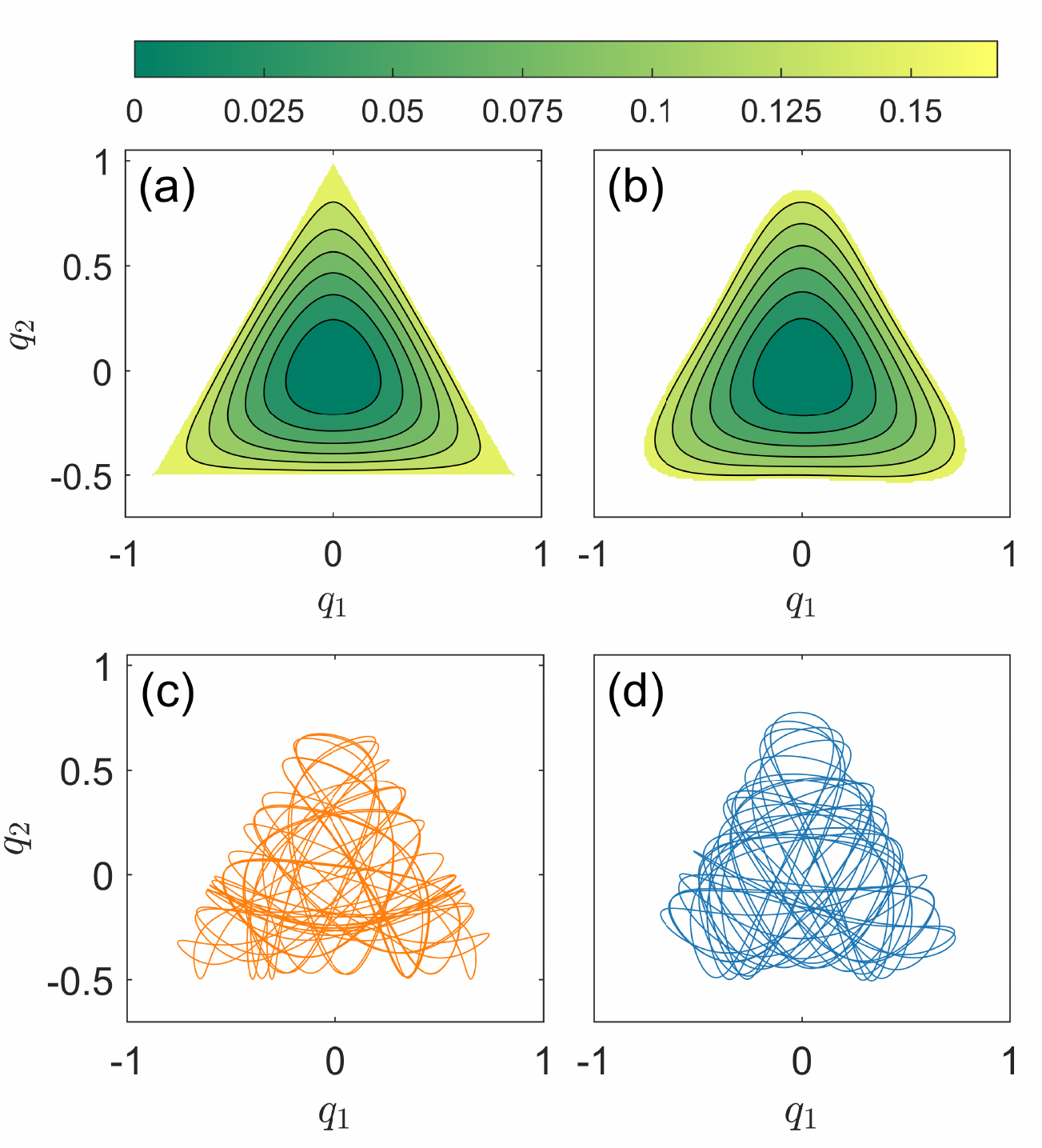}
\caption{Testing the adaptability of HNN for parameter values outside the
training interval. (a,b) True and predicted contour maps of the potential
function for $\alpha=1$. (c,d) True and predicted orbits from the same
initial condition $[q_1,q_2,p_1,p_2]=[0,0,1/\sqrt{6},1/\sqrt{6}]$, which
differ in detail but are both chaotic.}
\label{fig:HH_4}
\end{figure}

For $\alpha=1$, with energy $E=1/6$, most of the orbits are chaotic, where
the portion of the KAM islands in the phase space becomes relatively
insignificant, as shown in Fig.~\ref{fig:HH_1}(d). In the case, the contour map
of the true potential function has a triangular shape, as shown in
Fig.~\ref{fig:HH_4}(a). The predicted potential contour map is shown in
Fig.~\ref{fig:HH_4}(b), which agrees reasonably well with the true one.
Figures~\ref{fig:HH_4}(c) and \ref{fig:HH_4}(d) show a true and the
predicted chaotic orbits from the same initial condition. While their details
are different, the HNN predicts correctly that the orbit is chaotic.
In fact, as will be shown in Sec.~\ref{subsec:HH_chaos}, for the two quantities
characterizing the statistical behavior of the orbit, e.g, the maximum
Lyapunov exponent and the alignment index, the predicted orbit yields the same
results as those from the true orbit. In general, the closer the testing
parameter value is to one of the training points, the higher the prediction
accuracy.

\subsection{Adaptable prediction of a Hamiltonian system} \label{subsec:HH_chaos}

In a typical Hamiltonian system, the route of transition to ergodicity as
a nonlinearity parameter increases is as follows~\cite{LL:book}. In the weak
nonlinear regime, the system is integrable, where the motions are
quasiperiodic and occur on tori generated by different initial conditions,
as illustrated in Fig.~\ref{fig:HH_1}(a) for the H\'{e}non-Heiles system. As
the nonlinearity parameter $\alpha$ increases, chaotic seas of various sizes
emerge, leading to a mixed phase space, as exemplified in Fig.~\ref{fig:HH_1}(b).
In the regime of strong nonlinearity, e.g., $\alpha=1$, most of the phase
space constitutes chaotic seas with only a small fraction still occupied by
KAM islands, as shown in Fig.~\ref{fig:HH_1}(c) and (d). Here we provide strong evidence
for the adaptability of our parameter-cognizant HNN by demonstrating that it
can accurately predict the transition scenario, with training conducted based
on time series from only a handful values of the nonlinearity parameter.

Distinct from dissipative systems in which random initial conditions in the
basin of attraction of an attractor (periodic or chaotic) lead to trajectories
that all end up in the same attractor, in Hamiltonian systems different
initial conditions typically lead to different dynamically invariant sets.
Because of this feature of Hamiltonian systems, to investigate the transition
scenario, computations from initial conditions in the whole phase space
leading to a statistical assessment and characterization of the resulting
orbits are necessary. We focus on two statistical quantities: the largest
Lyapunov exponent and the minimum alignment index, where the former
characterizes the exponential separation rate of infinitesimally close
trajectories and the latter measures the relative ``closeness'' of two
arbitrary vectors along the trajectory (e.g., whether they become parallel,
anti-parallel, or neither)~\cite{skokos2001alignment}. For a chaotic
trajectory, an infinitesimal vector stretches or contracts exponentially
along the unstable or the stable direction, respectively. As a result, a
random vector will approach the unstable direction along the trajectory and
two random vectors will align with each other quickly. In particular, given
two initial vectors $\mathbf{u}^0_1$ and $\mathbf{u}^0_2$, after $i$ time
steps, they become $\mathbf{u}^i_1$ and $\mathbf{u}^i_2$, respectively. The
minimum alignment index is defined as
\begin{equation} \label{eq:alignment}
\gamma^i \equiv \text{min} (\| \mathbf{u}_1^i + \mathbf{u}_2^i\|, \| \mathbf{u}_1^i - \mathbf{u}_2^i\|).
\end{equation}
When chaos sets in, the value of $\gamma^i$ will quickly approach zero with
time.

\begin{figure} [ht!]
\centering
\includegraphics[width=\linewidth]{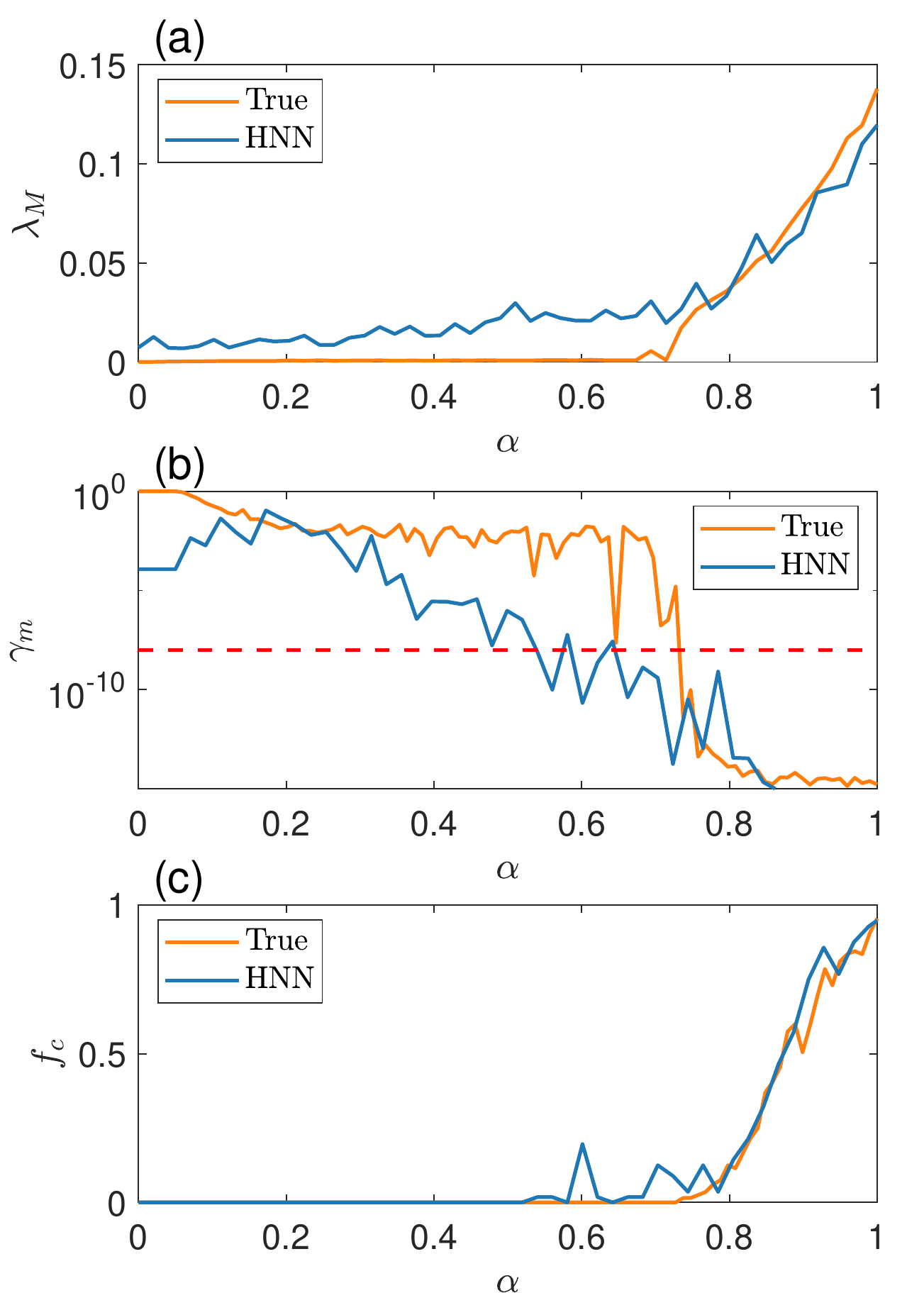}
\caption{ Test of adaptability of parameter-cognizant HNN in predicting
transition to chaos in the H\'{e}non-Heiles system. (a-c) The ensemble
maximum Lyapunov exponent $\lambda_M$, the ensemble minimum alignment index
$\gamma_m$ together with the threshold $10^{-8}$, and the fraction $f_c$ of
chaos in the phase space versus the nonlinearity parameter $\alpha$,
respectively. Transition to chaos occurs about $\alpha \agt 0.7$. The orange
and blue colors correspond to the true and HNN predicted results, respectively.
There is a reasonable agreement between the predicted and true behaviors,
attesting to the adaptable predictive power of the HNN.}
\label{fig:lambda}
\end{figure}

For a properly trained HNN with its weights and biases determined, the output
contains the predicted Hamiltonian whose partial derivatives with respect
to the coordinate and momentum vectors can be calculated directly based on
the architecture of the neural network. These partial derivatives constitute
the velocity field of the underlying dynamical system, whose Jacobian matrix
can then be determined, from which the machine predicted Lyapunov exponents
and the alignment index can then be calculated (see Appendix~\ref{Appendix}).
The true values of the Lyapunov exponents and the minimum alignment index can
be calculated directly from the original Hamiltonian~\eqref{eq:Hamiltonian}
of the target system.

In our calculation, we take 100 equally spaced values of the bifurcation
parameter in the unit interval: $\alpha \in [0,1]$. For each $\alpha$ value,
we choose 200 random initial conditions and calculate, for each initial
condition, the values of the largest Lyapunov exponent and the minimum
alignment index. A trajectory is deemed chaotic~\cite{zotos2015classifying}
if the largest exponent is positive and the minimum alignment index is less
than $10^{-8}$. We denote the maximum Lyapunov exponent and the minimum
alignment index from the ensemble of 200 trajectories as $\lambda_M$ and
$\gamma_m$, respectively, which are functions of $\alpha$. Another quantity
of interest is the fraction of chaotic trajectories, denoted as $f_c$, which
also depends on $\alpha$. The triplet of characterizing quantities,
$\lambda_M$, $\gamma_m$, and $f_c$, can be calculated from the HNN and from
the original Hamiltonian as a function of $\alpha$. A comparison can then be
made to assess the adaptable power of prediction of our parameter-cognizant HNN.

Figures~\ref{fig:lambda}(a-c) show the machine predicted and true values of
$\lambda_M$, $\gamma_m$, and $f_c$ versus $\alpha$, respectively, for particle
energy $E = 1/6$. It can be seen that chaos arises for $\alpha \agt 0.7$,
at which $\lambda_M$ becomes positive, $\gamma_m$ decreases to $10^{-8}$, and $f_c$
begins to increase from zero. In Fig.~\ref{fig:lambda}(a), the true value
of $\lambda_M$ for $0 \le \alpha < 0.7$ is essentially zero, but the HNN
predicted $\lambda_M$ is slightly positive. The remarkable feature is that
both types of $\lambda_M$ value begins to increase appreciably for
$\alpha > 0.7$. In fact, there is a reasonable agreement between the true
and predicted behavior of $\lambda_M$. Similar features are present in the
behaviors of $\gamma_m$ and $f_c$ versus $\alpha$, as shown in
Figs.~\ref{fig:lambda}(b) and \ref{fig:lambda}(c), respectively. These
results are strong evidence that our parameter-cognizant HNN is capable of
adaptable prediction of distinct dynamical behaviors in Hamiltonian systems.

\section{Issues pertinent to adaptability of Hamiltonian neural networks}\label{sec:issues}

We address the adaptability of HNNs by asking the following three questions.
First, can the adaptability of HNNs be enhanced by increasing the number of
training values of the bifurcation parameter? Second, can adaptability be
achieved with multiple parameter channels? Third, does adaptability hold
for different target Hamiltonian systems?

\subsection{Effect of number of training parameter values}

\begin{figure} [ht!]
\centering
\includegraphics[width=\linewidth]{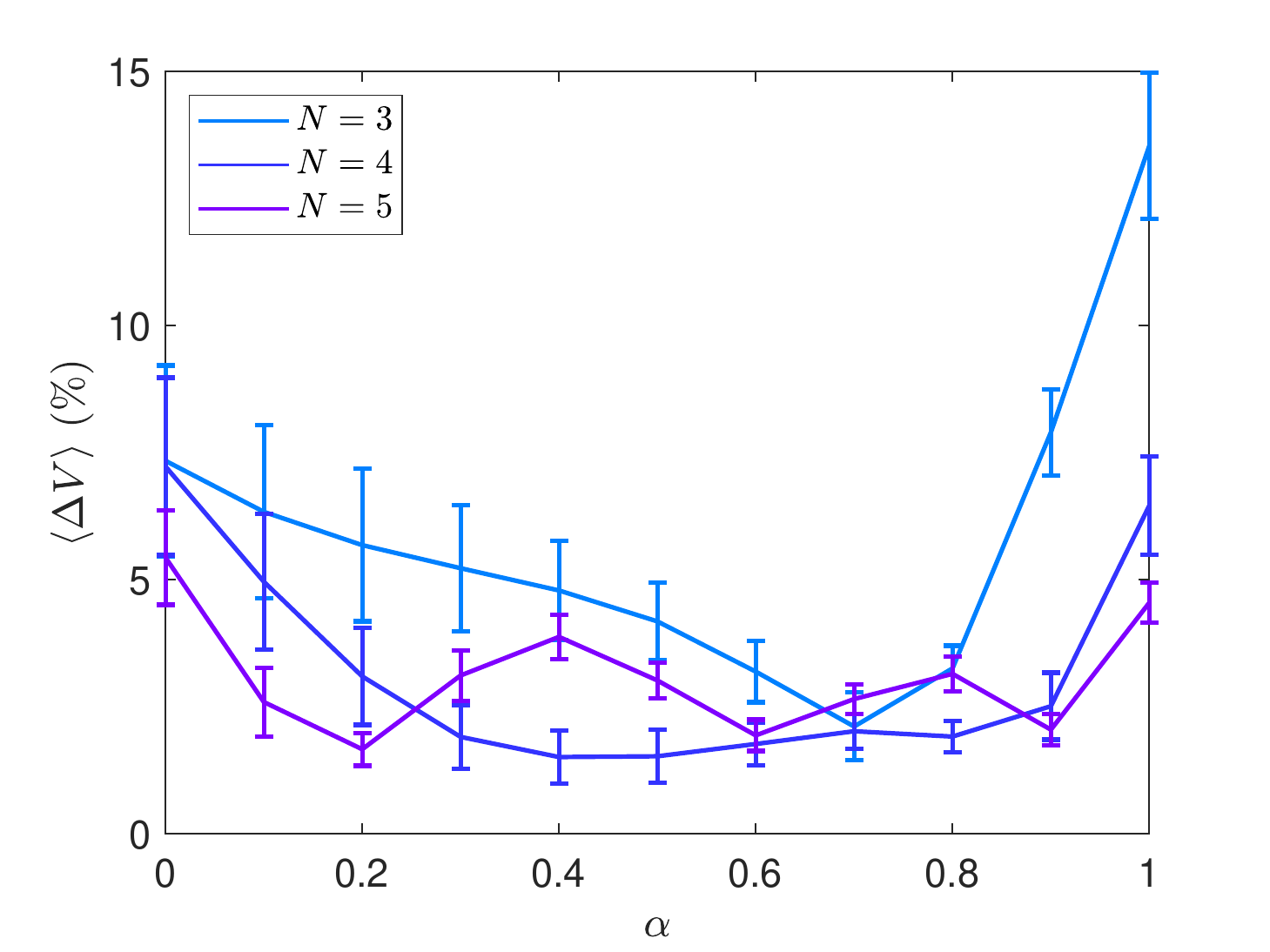}
\caption{Effect of increasing the number $N$ of training parameter values on
the adaptable prediction error. Shown are the ensemble errors
$\langle \Delta V\rangle$ in predicting the potential function for the
three simulation settings explained in the text. Increasing $N$ beyond four
does not lead to a significant reduction in the errors, indicating that
the HNN has already acquired the needed adaptability with training at four
different values of the bifurcation parameter.}
\label{fig:NM}
\end{figure}

So far, we have used four distinct values of the bifurcation parameter
to train our parameter-cognizant HNN. We now investigate if the adaptability
can be enhanced by increasing the number of training parameter values. Here
by ``enhancement'' we mean a reduction in the overall errors of predicting
the Hamiltonian in a parameter interval that contains values not in the
training set. To test this, we conduct the following numerical experiment.
We choose $N \ge 3$ training parameter values and, for each parameter value,
we train the HNN $M$ times using an ensemble of time series collected from $M$
energy values below the threshold (10 time series from 10 random initial
conditions with energy below the threshold). To make the comparison meaningful,
we choose the values of $M$ and $N$ such that $NM$ is approximately constant.
In particular, for Simulation $\#$1, we set $N = 3$: $\alpha = 0.25$, 0.5 and
0.75, and $M = 9$. For Simulation $\#$2, we choose $N = 4$: $\alpha = 0.2$,
0.4, 0.6, and 0.8, and $M = 7$. For Simulation $\#$3, we have $N = 5$:
$\alpha = 0.1$, 0.3, 0.5, 0.7, and 0.9, and $M = 5$. For each simulation, we
calculate the ensemble error $\langle \Delta V\rangle$ in predicting the
potential function as defined in Eq.~(\ref{eq:DV}) for $0 \le \alpha \le 1$.
The results are shown in Fig.~\ref{fig:NM}. It can be seen that the errors
for $N = 3$ are generally larger than those for $N > 3$, but the errors for
the two cases ($N=4$ and 5) are approximately the same, indicating that
increasing $N$ above four will not lead to a significant reduction in the
errors of adaptable prediction. That is, by training with multiple time series
from four values of the bifurcation parameter, the HNN has already acquired
the necessary adaptability for predicting the system behavior at other nearby
parameter values.

\subsection{HNNs with two parameter channels}

\begin{figure} [ht!]
\centering
\includegraphics[width=\linewidth]{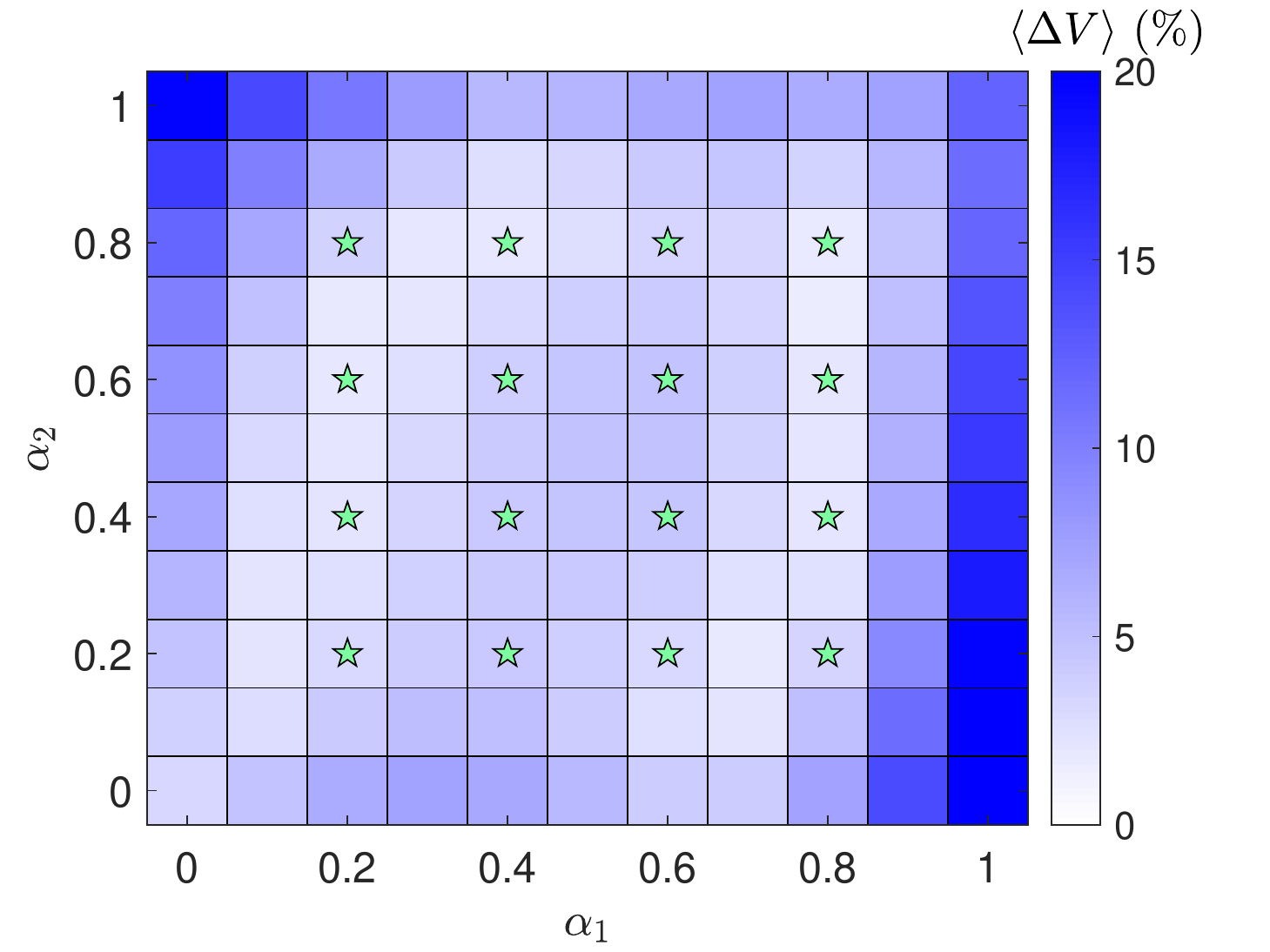}
\caption{Prediction performance of an HNN with two input parameter channels.
The target Hamiltonian system is given by the asymmetric H\'{e}non-Heiles
system as defined by Eq.~(\ref{eq:Hamiltonian_2}). Shown is the color-coded
ensemble prediction error $\langle \Delta V\rangle$ in the
$(\alpha_1,\alpha_2)$ plane. Training is conducted at the 16 points indicated
by the green pentagons. The prediction error is small ($< 5\%$) in the central
region $(\alpha_1,\alpha_2) \in [0.2, 0.8]$.}
\label{fig:2Channel}
\end{figure}

We construct parameter-cognizant HNNs with more than one parameter channel.
For this purpose, we modify the H\'{e}non-Heiles Hamiltonian
Eq.~\eqref{eq:Hamiltonian} to
\begin{equation} \label{eq:Hamiltonian_2}
H=\frac{1}{2}\left(p_1^2+p_2^2\right) + \frac{1}{2}\left(q_1^2+q_2^2 \right)+\alpha_1q_1^2q_2-\frac{\alpha_2}{3}q_2^3,
\end{equation}
where $\alpha_1$ and $\alpha_2$ are two independent bifurcation parameters,
requiring two independent parameter input channels to the HNN. The energy
threshold for bounded motions can be evaluated numerically. We conduct
training for a number of combinations of $\alpha_1$ and $\alpha_2$ values:
$\alpha_1,\alpha_2\in\{0.2, 0.4, 0.6, 0.8\}$. The training data are generated
as follows: for each parameter pair, we choose five energy values below the
threshold and, for each energy value, a single time series is collected.
After the training is done, we predict the potential function for
$\alpha_1,\alpha_2\in[0, 1]$ with the interval $0.1$ in each direction of
parameter variation. Figure~\ref{fig:2Channel} shows the color-coded ensemble
prediction error $\langle \Delta V\rangle$ in the $(\alpha_1,\alpha_2)$
plane. For some combinations of $\alpha_1$ and $\alpha_2$ with a relatively
large difference in their values, the threshold energy is less than $1/6$.
For such cases, the integration domain in Eq.~\eqref{eq:DV} is modified
accordingly based on the threshold value. It can be seen that, in the parameter
region $(\alpha_1,\alpha_2) \in(0.2, 0.8)$, the prediction error is about
$5\%$, while the errors outside the region tend to increase. At the two
off-diagonal corners, the errors are the largest, due to the strong asymmetry
in the potential profile. Figure~\ref{fig:2Channel} demonstrates that HNNs
with two parameter channels can be trained to be adaptable for prediction.

\subsection{HNNs for a diatomic molecule system}

\begin{figure} [ht!]
\centering
\includegraphics[width=\linewidth]{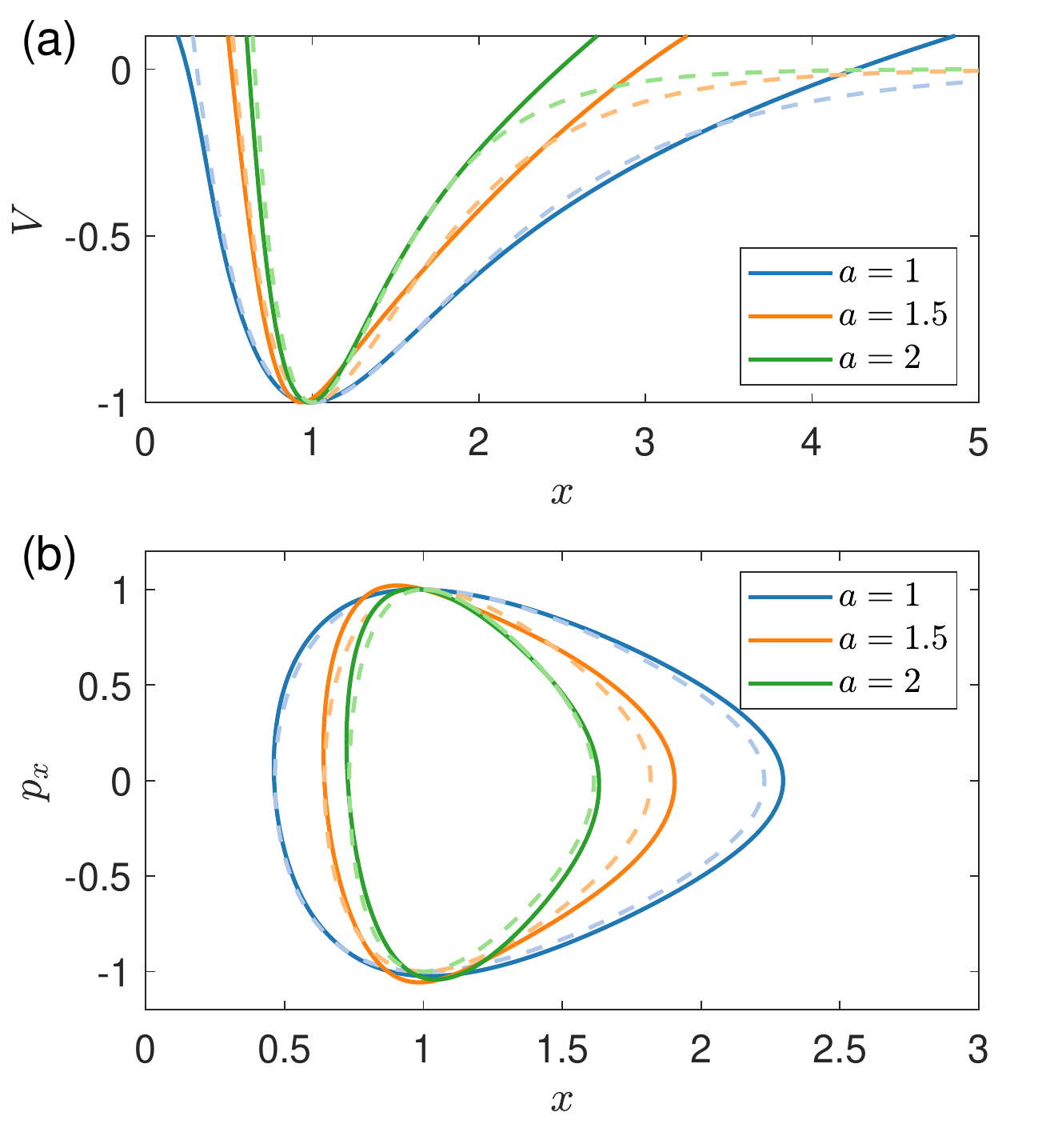}
\caption{Parameter-cognizant HNN trained for the one-dimensional Morse system.
(a) Shown are the predicted potential profiles for $a = 1.0$, 1.5, and 2.0
(solid curves), together with the corresponding true profiles (dashed curves).
The predicted potential function for $a = 1.5$ is not the interpolation of those
for $a = 1.0$ and $a = 2.0$, attesting to the adaptable predictive power of
the HNN. (b) True and predicted orbits in phase space from $[x,p_x]=[1,1]$.}
\label{fig:Morse}
\end{figure}

We consider a different target Hamiltonian system, a system defined by the
one-dimensional Morse potential that describes the interaction between
diatomic molecule~\cite{morse1929diatomic}. This system was previously
used in the study of chaotic scattering~\cite{Lai:2000,lai2000topology}.
The Hamiltonian is given by
\begin{equation}
	H=\frac{p_x^2}{2}+V(x) \equiv \frac{p_x^2}{2} + [1-\exp(-a(x-x_0))]^2-1,
\end{equation}
where the potential function $V(x)$ has a minimum value at $x=x_0$ with
$V(x_0)=-1$ and $V(x\rightarrow \infty)\rightarrow 0$. Taking the minimum
potential value as the reference point for energy $E$, all orbits are bounded
for $E < 1$. We set $x_0=1$ and choose $a$ as the bifurcation parameter.
The training data are generated from four different values of $a$: $a = 0.5$,
1.0, 2.0, and 4.0 where, for each training parameter value, an ensemble of
five values of energy is used, resulting an ensemble of 20 independent time
series. The time span for each time series is $0 \le t \le 100$ with the
sampling time step $\Delta t = 0.1$.

Figure~\ref{fig:Morse} shows the predicted potential profile for $a=1.5$,
together with those for the two training points $a=1$ and $a=2$. The result
is accurate for $x$ around the minimum potential point, but large errors
arise when the position is far away from the minimum point. A plausible
reason is that, for large values of $x$, the potential varies slowly, resulting
in small changes in the momentum. As a result, the corresponding portions of
the time series exhibit less variation, leading to large prediction errors by
the HNN. The trained HNN has apparently gained certain adaptability, as the
prediction result for $a = 1.5$ is not the interpolation of those for
$a = 1.0$ and $a = 2.0$.

\section{Discussion} \label{sec:discussion}

Developing adaptable machine learning in general has broad applications to
critical problems of current interest. For example, a problem of paramount
importance is to predict how a system may behave in the future when some
key parameters of the system may have drifted, based on information that
is available at the present. As an example, suppose an ecosystem is currently
in a normal state. Due to the environmental deterioration, some of its
parameters such as the carrying capacity and/or the species decay rates
will have drifted in the future. Is it possible to predict if the system
will collapse when certain amount of parameter drift has occurred, when
the system equations are not known and the only available information is
time series data that can be measured {\em prior to but including the
present}? Adaptable machine learning offers a possible solution. For example,
it has been demonstrated recently~\cite{KFGL:2021} that incorporating a
parameter-cognizant mechanism into reservoir computing machines enables
prediction of possible critical transition and system collapse in the future
for any given amount of parameter drift. However, the state-of-the-art
reservoir computing schemes under intensive current research~\cite{HSRFG:2015,LBMUCJ:2017,PLHGO:2017,LPHGBO:2017,DBN:book,LHO:2018,PWFCHGO:2018,PHGLO:2018,Carroll:2018,NS:2018,ZP:2018,WYGZS:2019,GPG:2019,JL:2019,VPHSGOK:2019,FJZWL:2020,ZJQL:2020} do not taken into account physical constraints such as energy
conservation, so they are suitable but for dissipative dynamical systems.

Combining the laws of physics and traditional machine learning has the
potential to significantly enhance the performance and predictive power of
neural networks. It has been demonstrated recently that enforcing the
Hamilton's equations of motion in the traditional feed-forward neural
networks can lead to improvement in the prediction accuracy for Hamiltonian
systems in both integrable and chaotic regimes~\cite{greydanus2019hamiltonian,
toth2019hamiltonian,bertalan2019learning,choudhary2019physics,
garg2019hamiltonian}. In these studies, training and prediction are conducted
for the same set of parameter values of the target Hamiltonian system, so the
underlying Hamiltonian neural networks are not adaptable in the sense that
they are not capable of predicting the dynamical behavior of the system at
a different parameter setting.

Do adaptable HNNs that we have developed have any practical significance? From
the point of view of making predictions of the future states of Hamiltonian
systems subject to parameter drifting, the answer is perhaps no. The main
reason is that HNNs require all coordinate and momentum time series. For
example, one may be interested in predicting whether a complicated many-body
astrophysical system may lose its stability and become mostly chaotic in the
future, where the only available information is the position and momentum
measurements prior to or at the present when the system is still in a mostly
integrable regime. As the laws of physics for this system are known, the data
required for training is not a lesser burden than knowing the Hamiltonian
itself. Nonetheless, our work generates insights into the working of HNNs,
as follows.

Our parameter-cognizant, adaptable HNNs have a parameter input channel to the
standard multilayer network with the loss function stipulated by the
Hamilton's equations of motion, and are capable of successful prediction of
transition to chaos in Hamiltonian systems. In particular, through training
with coordinate and momentum time series from four different values of the
bifurcation (nonlinearity) parameter, the machine gains adaptability as
evidenced by its successful prediction of the dynamical behavior of the target
system in an entire parameter interval containing the training parameter
values. That is, the benefits of training are that the HNN has learned not
only the dynamical ``climate'' of the target Hamiltonian system but also
how the ``climate'' changes with the bifurcation parameter. Machine learning
can thus be viewed as a process by which the neural network self-adjusts its
dynamical evolution rules to incorporate those of the target system.

When systematically varying values of the bifurcation parameter
are fed into the HNN, it can predict the transition to chaos from a mostly
integrable regime, as determined by the ensemble maximum Lyapunov exponent
and minimum alignment index as well as the fraction of chaos as a function
of the bifurcation parameter. For a single parameter channel, the adaptable
predictive power is achieved insofar as the training parameter set contains
at least three or four distinct values. For an HNN with duplex parameter
channels, the size of the required training parameter set should be at least
four by four. Adaptable prediction has also been accomplished for a different
Hamiltonian system defined by the Morse potential function. We expect the
principle of designing parameter-cognizant HNNs and the training method
devised in this paper to hold for general Hamiltonian systems.

One issue is the dependence of the energy surface on the bifurcation parameter.
As the parameter changes continuously, the energy surface will evolve
accordingly. If we intend to predict the system dynamics for some specific
value of the bifurcation parameter for a fixed energy value, the training
data sets should contain time series collected from a larger energy value
to cover the pertinent phase space region at the desired energy value.

It should also be noted that, using HNNs to predict the transition from
integrable dynamics to chaos in the H\'{e}non-Heiles system was first
reported~\cite{choudhary2019physics}, which relied on using energy $E$ as the
control parameter for a fixed value of the nonlinearity parameter
(e.g., $\alpha = 1$). Here we have studied the transition using $\alpha$ as
the bifurcation parameter for a fixed energy value (e.g., $E = 1/6$). The two
routes are equivalent because the H\'{e}non-Heiles system
Eq.~\eqref{eq:Hamiltonian} possesses a three-fold symmetry in the configuration
space. Such an equivalence also arises in systems whose potential function
contains nonlinear square terms, e.g., the classical $\phi^4$ or FPU
model~\cite{fermi1955studies,caiani1998geometry}. However, for the
two-parameters Hamiltonian Eq.~\eqref{eq:Hamiltonian_2} studied in this
paper, the three-fold symmetry is broken, destroying the equivalence between
varying the nonlinearity parameter and energy. In fact, for Hamiltonian
systems such as the Morse and double-pendulum systems, the equivalence
does not hold either~\cite{morse1929diatomic,lai2000topology,levien1993double}.
Our adaptable HNN does not rely on any such equivalence, and can be effective
in predicting the transition to chaos in any type of Hamiltonian systems.

\section*{Acknowledgment}

We would like to acknowledge support from the Vannevar Bush Faculty
Fellowship program sponsored by the Basic Research Office of the Assistant
Secretary of Defense for Research and Engineering and funded by the Office
of Naval Research through Grant No.~N00014-16-1-2828.

\appendix

\section{Algorithm for calculating the Lyapunov exponent and alignment index of
Hamiltonian neural networks} \label{Appendix}

Given a dynamical system $d\mathbf{x}/dt=\mathbf{f}(\mathbf{x})$, the
Jacobi matrix is given by $\mathcal{J}=\partial\mathbf{f}/\partial\mathbf{x}$.
For a Hamiltonian system, the dynamical variables are
$\mathbf{x} \equiv [\mathbf{q},\mathbf{p}]^T$ and
\begin{equation}
\mathbf{f}(\mathbf{q},\mathbf{p})=\left[\frac{\partial H}{\partial \mathbf{p}}, -\frac{\partial H}{\partial\mathbf{q}}\right].
\end{equation}
For an HNN, in principle, the Hamiltonian $H$ is given by a sequence of
operations of the neural network with the weights and biases in
Eq.~\eqref{eq:1_layer} determined by training. An alternative but efficient
approach to calculating the Jacobian matrix $\mathcal{J}$ is the
finite-difference method. In particular, for a given initial condition, we
generate an orbit of $N$ points with time interval $dt$ and calculate
$\mathcal{J}$ at each time step. Let the sequence of Jacobian matrices be
denoted as $\mathcal{J}(t_0), \mathcal{J}(t_1), \cdots, \mathcal{J}(t_N)$,
and let $\mathcal{Y}(t_0)$ be the identity matrix $\mathcal{I}$. If the phase
space of the target Hamiltonian system is $D$-dimensional ($D = 4$ for the
H\'{e}non-Heiles system), there are $D$ Lyapunov exponents. Let
$\bm{\lambda}$ be the vector of the $D$ Lyapunov exponents:
$\bm{\lambda} \equiv (\lambda_1,\lambda_2,\ldots,\lambda_D)^T$ and set the
initial values of the exponents to be zero:
$\bm{\lambda}(t_0) = (0,0,\ldots,0)^T$. After $N$ steps, we have
\begin{equation}
\mathcal{Y}(t_N) = \mathcal{J}(t_N) \cdot \mathcal{Y}(t_{N-1}).
\end{equation}

Carrying out the QR decomposition of the matrix $\mathcal{Y}(t_N)$ with the
resulting matrices denoted as $\mathcal{Q}$ and $\mathcal{R}$, we have
\begin{eqnarray}
\lambda_j (t_N) & = & \lambda_j (t_{N-1}) + \log{|R_{jj}|}, \\
\mathcal{Y} (t_N) & = & \mathcal{Q},
\end{eqnarray}
where $R_{jj}$ is the $j$th diagonal element of the matrix $\mathcal{R}$.
The Lyapunov exponents are given by
\begin{equation}
\lambda_j = \lim_{N\rightarrow\infty} \frac{\lambda_j (t_N)}{Ndt}, \ j = 1, \ldots, D.
\end{equation}
The maximum Lyapunov exponent is
$\lambda_M=\text{max}_j (\lambda_j)$.

To calculate the alignment index, we introduce a matrix $\mathcal{M}$ and set
it to be the identity matrix at the initial time:
$\mathcal{M} (t_0) = \mathcal{I}$.
Let $\mathbf{u}_1(t_0)=[1,0,0,0]^T$ and $\mathbf{u}_2(t_0)=[0,1,0,0]^T$ be two
linearly independent vectors at the initial time. After $N$ steps, we have
\begin{equation}
\begin{split}
\mathcal{M} (t_N) & =(\mathcal{I}+ \mathcal{J}(t_{N})dt) \cdot \mathcal{M}(t_{N-1}), \\
\mathbf{u}_{1,2} (t_N) & = \mathcal{M} (t_N) \cdot \mathbf{u}_{1,2}(t_0).
\end{split}
\end{equation}
Normalizing the vectors $\mathbf{u}_{1,2}(t_N)$ by their respective magnitude
to have the unit length, we obtain the minimum alignment index as
\begin{equation}
\gamma_m = \lim_{N\rightarrow \infty} \text{min} (\| \mathbf{u}_1 (t_N) + \mathbf{u}_2 (t_N)\|, \| \mathbf{u}_1 (t_N) - \mathbf{u}_2 (t_N) \|).
\end{equation}


%
\end{document}